\documentclass[
]{ceurart}

\usepackage{listings}
\begin{document}

\copyrightyear{2022}
\copyrightclause{Copyright for this paper by its authors.
  Use permitted under Creative Commons License Attribution 4.0
  International (CC BY 4.0).}

\conference{Woodstock'22: Symposium on the irreproducible science,
  June 07--11, 2022, Woodstock, NY}

\title{VAR: Visual Analysis for Rashomon Set of Machine Learning Models' Performance}

\author[1]{Yuanzhe Jin}[
]
\cormark[1]
\address[1]{University of Oxford, United Kingdom}

\cortext[1]{Corresponding author.}

\begin{abstract}
Evaluating the performance of closely matched machine learning(ML) models under specific conditions has long been a focus of researchers in the field of machine learning. The Rashomon set is a collection of closely matched ML models, encompassing a wide range of models with similar accuracies but different structures. Traditionally, the analysis of these sets has focused on vertical structural analysis, which involves comparing the corresponding features at various levels within the ML models. However, there has been a lack of effective visualization methods for horizontally comparing multiple models with specific features. We propose the VAR visualization solution. VAR uses visualization to perform comparisons of ML models within the Rashomon set. This solution combines heatmaps and scatter plots to facilitate the comparison. With the help of VAR, ML model developers can identify the optimal model under specific conditions and better understand the Rashomon set's overall characteristics.

\end{abstract}


\maketitle



\section{Introduction\label{sec:Intro}}

In the field of machine learning (ML), there has been a growing interest among researchers in conducting more nuanced evaluations of model performance under specific conditions. ML developers have a strong need to understand the similarities and differences among various ML models, especially when their performance metrics (such as accuracy or F1 score) are nearly identical. This challenge is encapsulated in the concept of the "Rashomon set," which refers to a collection of models that exhibit similar performance but differ in their structural properties. The existence of Rashomon sets implies that multiple models can achieve comparable accuracy while employing different strategies and structures to process data~\cite{breiman2001statistical}.


Historically, the analysis of models within Rashomon sets has focused on vertical structural analysis. In the context of decision tree models, vertical structural analysis involves comparing structural differences between decision trees of varying depths to analyze the model's decision-making process. This approach entails comparing the corresponding features used at different depths~\cite{wang2022timbertrek}. However, this method has limitations in terms of horizontal comparisons, which aim to analyze the specific features utilized across models. This deficiency becomes worse when ML developers and researchers need to compare and filter large numbers of similar models. 


To assist ML model developers in overcoming this limitation, we introduce VAR, a visual analysis for the Rashomon set of machine learning models as a solution. VAR is a visual analysis application that leverages visualization methods to analyze multiple ML model performances. It employs mathematical techniques based on radial basis functions to facilitate horizontal comparisons of ML models within Rashomon sets. By combining heatmaps and scatter plots, VAR provides a visual representation for the comparison of numerous models, enabling developers to effectively compare specific features across multiple models. In this paper, we present the development process and application of VAR, demonstrating its effectiveness through case studies and user evaluation.


The significance of this research lies in addressing a gap in ML model evaluation by enabling systematic horizontal comparisons within Rashomon sets. This empowers developers to gain deeper insights into model behaviors, leading to improved decision-making when selecting deployment models. For instance, in financial contexts (e.g., FICO dataset), VAR can assist in selecting robust loan approval models by comparing their decision-making strategies under similar performance metrics. Similarly, in law enforcement, it can help analyze crime prediction models derived from police databases (e.g,. COMPAS dataset), ensuring that selected models are not only accurate but also fair and interpretable. Furthermore, VAR contributes to the broader field of interpretable ML by offering tools that analyze model diversity and robustness more accessible, especially under challenging scenarios such as missing data. By streamlining this complex analysis, VAR has the potential to advance the development of reliable and fair ML systems across various domains.


The main contributions of VAR are in three parts:
\begin{enumerate}

\item VAR provides a comprehensive understanding tool. VAR helps ML model developers enhance the understanding of overall characteristics within Rashomon sets, providing insights into the diversity and robustness of models within the set. 

\item VAR offers an interactive platform. ML model developers can utilize VAR to identify issues in the collection of Rashomon sets, thereby improving the collection process.

\item VAR delivers an analysis and evaluation framework. VAR assists ML model developers in analyzing and determining the optimal model. It helps evaluate the performance of different ML models under certain missing data scenarios.

\end{enumerate}




We address a gap in current ML evaluation practices by introducing a visual analysis process and presenting an intuitive tool for horizontal exploration of ML models in the Rashomon Set. This visualization framework enhances the interpretability and transparency of the model selection process, facilitating a more nuanced understanding of model diversity and decision logic.

\section{Related Work}
\label{sec:RelatedWork}
\sloppy
\subsection{Rashomon Set in ML}

The idea of the Rashomon set was first discussed by Leo Breiman~\cite{breiman2001statistical}, who noted that many distinct models could achieve similar performance on the same task. Since this phenomenon was first noted, a substantial body of work studying the Rashomon set — the set of near-optimal models for a problem — has developed. Theoretical work has noted that noise in a dataset leads to the existence of large Rashomon sets~\cite{semenova2022existence, semenova2024path} and that working with finite data makes it inevitable that multiple indistinguishably “good” models will exist~\cite{paes2023inevitability}. This phenomenon, also referred to as “predictive multiplicity,” underscores the challenge of relying on a single model to encapsulate the underlying patterns of a dataset. Predictive multiplicity has been quantified and shown to be prevalent in multiple classification settings~\cite{marx2020predictive, watson2023predictive, hsu2022rashomon}, and has implications for fairness and accountability in machine learning~\cite{sahoh:2023:role}.


Recently it has become possible to completely enumerate Rashomon sets for decision trees~\cite{xin2022exploring} and generalized additive models~\cite{zhong2024exploring}, while other work computes incomplete samplings from Rashomon sets for rule lists~\cite{ciaperoni2024efficient} and neural networks. These advances allow researchers and practitioners to study how different models within the Rashomon set differ in structure, feature importance, and predictions~\cite{liu:2024:faim}. Further, they have enabled the exploration of diverse applications, including fairness auditing~\cite{coston:2021:characterizing} and causal inference, where different models provide complementary insights about causal relationships while remaining consistent with the observed data. The phenomenon of large Rashomon sets has also been demonstrated to arise in noisy and dynamic industrial settings, where studying these sets provides insights into model robustness over time~\cite{watson2024predictive}.


The existence of multiple valid models within a Rashomon set complicates traditional approaches to assessing feature importance, which typically rely on a single model. Several works have challenged this practice by studying variable importance across all models in the Rashomon set~\cite{fisher2019all, donnelly2023rashomon, smith2020model, dong2019variable}. These studies highlight the risks of drawing misleading conclusions from single-model analysis, emphasizing the importance of using Rashomon sets for more robust and transparent model interpretation.


Most relevant to this work, a handful of tools for visually exploring Rashomon sets have been developed. TimberTrek~\cite{wang2022timbertrek} introduced a sunburst-style interface for exploring the Rashomon set of decision trees, and GamChanger~\cite{wang2021gam} provides an interactive interface for generalized additive models. Both tools help users evaluate trade-offs between model simplicity, fairness, and interpretability. One similar to our work~\cite{dong2019variable} suggested creating scatter plots where each axis represented the importance of a given variable, and each dot corresponded to a single model. This approach, while insightful, was not designed for broader accessibility or for exploring relationships beyond variable importance. Recent advancements in tools for Rashomon set analysis promise to improve the ability of practitioners to engage with these ideas~\cite{biecek:2024:performance, rudin2024amazing}.


For a comprehensive review of Rashomon sets, including theoretical foundations, practical implications, and emerging tools, see~\cite{rudin2024amazing}, which provides a detailed overview of this critical concept in modern machine learning.


\subsection{Visual Analysis for ML Models}

Visual analysis techniques for ML models have evolved from simple data visualizations like scatter plots and bar graphs to more advanced methods for visualizing model features. Interactive tools now enable real-time parameter adjustments, improving model interpretability, tuning, and diagnostics.


Researchers have highlighted the value of visualization across different ML stages, as seen in the knowledge graph of ML visualization by Sacha et al.~\cite{Sacha:2018:TVCG}, and a hypothesis testing method for feature learning by Chatzimparmpas et al.~\cite{chatzimparmpas2020state}. Visualization techniques have supported the analysis and understanding of a wide range of ML models, from basic algorithms to deep learning. For instance, CNNs have been extensively studied through visualization~\cite{Sietzen:2021:CGF}, as have RNNs~\cite{Shen:2020:PVIS}. Traditional methods like decision trees~\cite{Streeb:2021:TVCG} and random forests~\cite{Gurung:2019:ECDA} have also benefited from visualization research. Additionally, reinforcement learning has seen advancements in visualization applications~\cite{Wang:2021:TVCG}, demonstrating the broad potential of visualization in enhancing the explanation, optimization, and understanding of ML algorithms.


Visualization-assisted techniques have been used to develop ML models for various applications, such as in the field of music analysis~\cite{Ye:2022:TVCG}, text analysis~\cite{jin:2024:igaiva}, weather forecasting~\cite{Palaniyappan:2022:aqx}, and image classification~\cite{Wang:2020:TVCG}. These application cases not only cover traditional data analysis domains but also extend to more complex system control and prediction tasks, highlighting the important role of visualization for ML workflow in the development and application of ML models.


Visualization techniques effectively support the analysis of diverse ML models. Since Rashomon sets arise from various application scenarios, leveraging visualization can assist ML developers in analyzing and filtering these sets. Our work aims to provide valuable tools to help researchers better analyze and utilize their model sets.


\subsection{Decision Trees}

As a widely utilized ML model, a decision tree is known for its interpretability~\cite{Quinlan:1986:kap}. Feature selection in decision trees also aids researchers in identifying more effective feature subsets for specific problems.

The foundational concept of recursive partitioning was initially proposed by early computational learning theorists, with contributions from seminal works~\cite{Breiman:1984:book}. Researchers developed techniques like information gain~\cite{Quinlan:1986:kap}, gain ratio~\cite{Salzberg:ML:1994}, and impurity measurements~\cite{Lewis:2000:book}. Some methods, such as GBDT~\cite{Ye:2009:ACM}, LightGBM~\cite{Ke:2017:NIPS}, and XGBoost tree~\cite{Chen:2016:KDD} have been introduced to help the decision tree construction process. They use different gradient boosting techniques to improve the efficiency and accuracy of ML.


In this paper, we use the generalized optimal sparse decision tree~\cite{hayden:2022:fast} to train our ML models. Our approach assumes that VAR can be implemented across different tree-based models for performance analysis. Utilizing a decision tree model enables us to extract and understand the internal model structures and feature interactions.

\section{VAR: Background, Visualization, and System}
\label{VAR_system}
To begin with, it is important to define the concept of a Rashomon set in the context of ML models. A Rashomon set refers to a collection of models that achieve similar levels of accuracy but offer different interpretations or decisions when applied to the same dataset. Studying Rashomon sets serves several purposes, including but not limited to improving model interpretability, addressing issues related to bias and fairness, and analyzing the advantages of ensemble methods. By exploring the Rashomon set, we can gain a deeper understanding of how different models interpret the same data, assess the robustness of these interpretations, and investigate strategies for selecting models when their performance is comparable.


\subsection{Background}
\label{sec:Background}

Though the Rashomon set has various applications, through interviews with ML developers from D University, we gained insight into a certain issue currently faced by developers.
Given a training dataset without missing data, ML developers find a Rashomon set of near-optimal models. The optimality is measured based on the regularized training loss across all samples in the given dataset. At a high level, ML developers want to understand how different models in this Rashomon set perform according to another metric, such as performance on a subset of the data, or whether they violate certain constraints. The idea is to understand performance on requirements that were not explicitly optimized for during training. One thing that needs to be made clear is that missing data is different from small datasets. There are many reasons for missing data. Feature values may be missing due to incomplete data collection or measurement limitations. The final result is that in a dataset, some data items cannot participate in training or decision-making when the model is trained.


In a setting with missing data, a given model may be invalid for certain samples because the model might depend on unknown features. The model's performance on subsets with specific missing patterns may differ from the overall performance. ML model developers also want to leverage the fact that different missing patterns can have related structures: some samples might be missing both feature A and feature B, while others might be missing exactly one of these two features. ML model developers desire visualization capabilities to help understand which models might be most suitable for different sub-structures of missing patterns.


Regarding the current solutions to this issue, through continued interviews with ML model developers, we found that many respondents choose different methods based on the type and proportion of missing data. For data that is missing at random, they tend to use statistical methods for imputation; whereas for data that is missing not at random, they are more inclined to analyze the reasons for the missing data and try to address the issue through additional data collection or feature engineering methods~\cite{Zheng:2018:book}.


Besides, we collected the requests that most model developers generally believe that transparency and interpretability are essential when handling missing data. They emphasized the need to handle missing data to ensure the credibility and reproducibility of the model.  We also noted that developers are interested in using visualization techniques to help identify patterns of missing data, thereby formulating more targeted screening and classification of models. Although some past tools~\cite{dong2019variable} could help address this dilemma to a certain extent, they lacked intuitive visualizations as support.


Based on the organization of the interview content above, we summarize the difficulties in helping ML model developers screen suitable models from the Rashomon set into the following three questions:


\begin{enumerate}
\item \textbf{Q1}: How to analyze the relationship between performance and features of multiple ML models in the Rashomon set?

\item \textbf{Q2}: How to visualize the performance and feature importance of multiple ML models in the Rashomon set?

\item \textbf{Q3}: How can ML model developers with an effective tool that allows them to conveniently collect and screen different ML models in the Rashomon set?

\end{enumerate}



\subsection{Radial Basis Functions for Visualization}

For \textbf{Q1}, regarding multiple models performances data as multiple data points, we drew inspiration from a previous paper~\cite{jin:2024:igaiva}, realizing that radial basis functions are not only suitable for analyzing high-dimensional data after projection but also serve a similar purpose in evaluating decision tree-based models across multiple different feature dimensions. For analyzing relationships between multiple models, radial basis functions can provide a relatively intuitive comparison of multiple models helping ML developers quickly screen out suitable target models.


Radial basis functions (RBFs) are a class of mathematical functions widely used for function approximation and pattern recognition in ML and numerical analysis. 
Given an input vector $\mathbf{x} \in \mathbb{R}^d$ and a set of center points $\{\mathbf{c}_i\}_{i=1}^N$, the general form of an RBF is defined as:
\[
\phi(r) = \phi(\|\mathbf{x} - \mathbf{c}_i\|), \quad r = \|\mathbf{x} - \mathbf{c}_i\|,
\]
where $\|\mathbf{x} - \mathbf{c}_i\|$ is the Euclidean distance between $\mathbf{x}$ and $\mathbf{c}_i$. In this work, we adopt a specific kernel function expressed as:
\[
\phi(r) = \frac{r \cdot \log(1 + r^{0.5})}{1 + r^{0.1}},
\]
which combines logarithmic and smoothing terms. The term $\log(1 + r^{0.5})$ enhances the local feature capture for small distances, while $(1 + r^{0.1})^{-1}$ ensures global stability for large distances. This design allows the function to maintain a balance between local adaptability and global smoothness.

By applying the RBF interpolation, we map model performance across various features into a continuous space, which can then be visualized using heatmaps and scatter plots. This approach facilitates the intuitive comparison of multiple ML models under varying feature conditions, providing deeper insights into the Rashomon set's diversity and robustness.


If we create a set from a model's performance across various features and input the corresponding parameter results as a vector into the general form of RBF, we can obtain a set of RBF values for the model. After performing the same operation for models to be compared in the Rashomon set, we generate a figure composed of RBF values to visually compare different ML models, thereby helping to answer \textbf{Q1}.


\begin{figure}[!ht]
  \centering
  \includegraphics[scale=0.3]{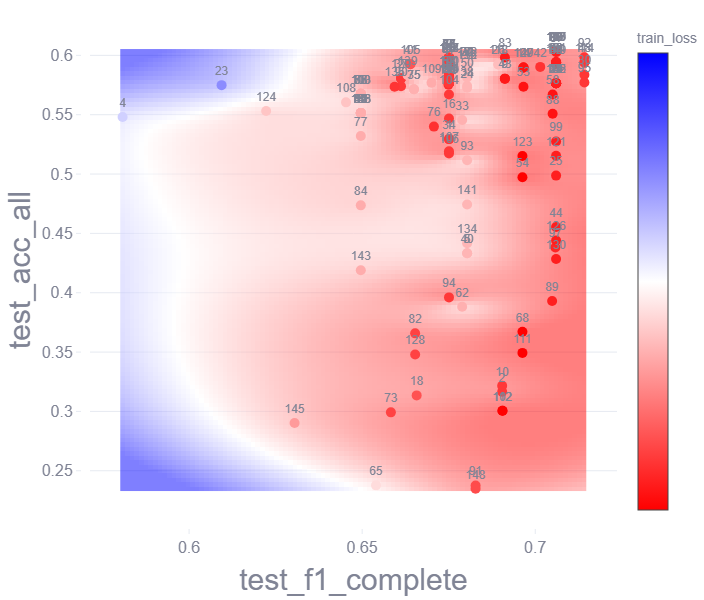}
  \includegraphics[scale=0.3]{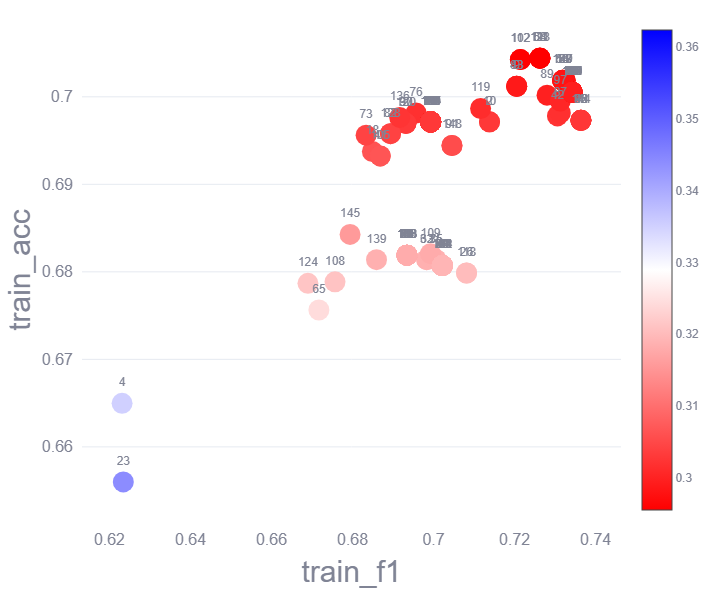}
  \caption{(a) RBF-heatmap mode (Left). Comparison of the performance of 152 models in the Rashomon set on the test set. To ensure that the color of the dots does not completely blend with the background, the color of the dots has been darkened. The color represents the train loss. (b) RBF-dot mode (Right). Comparison of the performance of the same 152 models on the training set. The color represents the train loss.}
\label{rbf_mode}
\end{figure}

\subsection{Visual Design of Radial Basis Functions}

One approach to visualizing the Rashomon set that has gained traction is scatter plot-based analysis, as suggested by~\cite{dong2019variable}. In this approach, each axis represents the importance of a specific variable, and each point corresponds to a model within the Rashomon set. This method allows users to explore relationships between variables across different models, offering valuable insights into how variable interactions might impact model predictions. 


Upon realizing that using RBF can analyze the relationships between the performance and features of multiple ML models in the Rashomon set, we naturally begin to consider the appropriate visualization methods to aid ML model developers. Through discussions with ML model development experts, the first concern mentioned was that simply using scatter plots is often not intuitive enough. After comparing scatter plots and heatmaps, we introduced two visualization modes: \textbf{RBF-heatmap} mode and \textbf{RBF-dot} mode.


\textbf{RBF-heatmap} mode focuses on providing an intuitive understanding of the performance of a large number of models. Its visualization design draws inspiration from the design in Jin's work~\cite{jin:2024:igaiva}. Fig.~\ref{rbf_mode}(a) shows the comparison of models collected in the Rashomon set under the \textbf{RBF-heatmap} mode. In the figure, the red and blue colors provide an intuitive representation of the model performances under certain constraints in the two testing metrics.


\textbf{RBF-dot} mode builds upon \textbf{RBF-heatmap} and incorporates user feedback, focusing primarily on filtering and comparing existing model results. In Fig.~\ref{rbf_mode}(b), we use colors to paint the dots, with the dot's color calculated by the RBF based on its position. By enlarging the dots representing each model, this helps ML model developers focus on comparing models based on two existing features (x, and y-axis) while comparing a third type of feature. 


The development of the \textbf{RBF-dot} mode was not accomplished at the beginning. Through discussion with ML model developers, the design was continuously improved. This development process is inspired by Ben Shneiderman's work~\cite{Shneiderman:2022:human}, which emphasizes the importance of close communication with users during tool development and improving the design through an iterative process.


To help ML model developers better analyze and filter models in the Rashomon set, we adjusted the size of the dots in the scatter plot. By controlling the size of the dots, we shifted the visualization focus from coloring the surrounding regions to visualizing the dots themselves, thereby addressing the ML model developers' concerns regarding interpretability. 


While both visualization modes rely on RBF interpolation, they serve different purposes and target distinct analytical scenarios. The RBF-heatmap mode uses a continuous color gradient to encode interpolated performance values across the feature space. By leveraging interpolation, the heatmap mode offers a global perspective, allowing users to identify broad patterns, such as clusters of high-performing models or regions of feature combinations where performance varies. 
In contrast, the RBF-dot mode focuses on individual models, visualized as discrete scatter points within the feature space. Each point corresponds to a specific model, with its position determined by two selected feature dimensions. The color of each point encodes the same performance metric as in the heatmap mode, providing a direct comparison of model-specific performance. Unlike the heatmap mode, this approach avoids interpolation, making it more precise for examining actual data points. The dot size can be adjusted to emphasize specific models or highlight their importance within the Rashomon set. This mode is suited for tasks that require detailed, model-level analysis, such as identifying top-performing models, exploring outliers, or evaluating performance under specific feature conditions. We present the case study of the two modes in Section~\ref{sec:case_study}. Given the two modes of the RBF-based visualization methods, we propose a solution to \textbf{Q2}.


\subsection{VAR System}

\begin{figure}[!ht]
  \centering
  \includegraphics[scale=0.6]{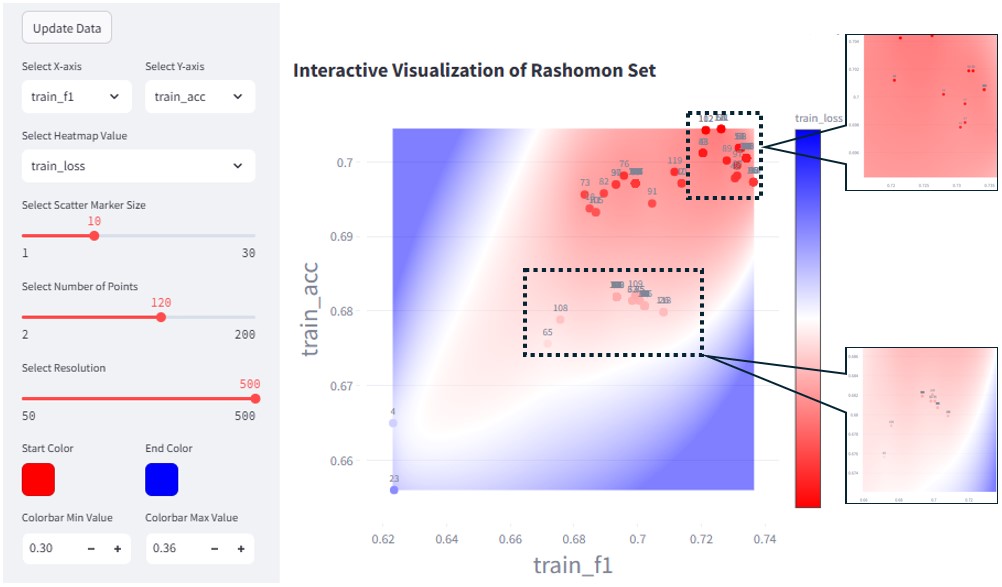}
  \caption{A screenshot of VAR showing functions in the control panel on the left and an RBF visualization plot on the right.}
\label{VAR_UI}
\end{figure}

After addressing Q1 and Q2, a natural demand arises from ML model developers: how to have an effective tool that allows filtering models in the Rashomon set using RBF-based visualization methods. This brings us to the resolution of the Q3.


To this end, we developed the VAR. VAR is a web-based visual analytics application. To facilitate easy operation for ML model developers, the design of VAR was inspired by Ben Shneiderman's design principles. Shneiderman proposed the visualization design principle of "overview first, zoom and filter, then details on demand”~\cite{Shneiderman:2000:book}. The web interface is primarily divided into a control panel on the left and a display area on the right, as shown in Fig.~\ref{VAR_UI}.


The control panel offers a wide range of features. As shown in Fig.~\ref{VAR_UI}, in the data section, users can not only replace the data via the "Update Data" button but also set corresponding features or input data parameters for the three visualization dimensions (x, y, colorbar). Below the input data controls, the visualization features are displayed, from controlling the scatter marker size to adjusting the number of data points displayed to set the image resolution. There is also great flexibility in selecting different colors for the colorbar and setting the data range mapped to the colors. Based on user feedback, we made targeted improvements to this section to make it more adaptable to different needs.


The development of VAR took a total of 9 months, during which the VAR developer (the first author) and the ML model developers held regular online meetings, totaling 18 meetings. The first author also participated in larger meetings held by the ML model developers and their collaborators, where they presented the design and applications of VAR in the conference room. The ML model developers assisted the first author in collecting user feedback. Based on the feedback, the design of VAR evolved from a localized version to a web-based version. Through ongoing communication among the developers, the design of VAR continued to improve, and the RBF-heatmap mode was developed into the RBF-dot mode. Additionally, with the assistance of a researcher from Z Institute, the first author optimized the color functionalities. The VAR system has now undergone a small-scale test at the university where the ML model developers are based, and feedback collected from these tests is presented in Section~\ref{sec:evaluation}. At this point, through the development of VAR, we have addressed \textbf{Q3} mentioned in Section \ref{sec:Background}, and have preliminarily verified the effectiveness of VAR through the results presented in the subsequent user evaluation section. 

\section{Case Study}
\label{sec:case_study}

Besides specific tasks from individual experts, we demonstrate the effectiveness of VAR through experiments on two publicly available datasets. The Rashomon sets were collected from the FICO dataset and the COMPAS dataset, respectively. These two datasets were chosen to evaluate VAR across different domains of datasets.


\begin{figure}[!ht]
  \centering
  \includegraphics[scale=0.3]{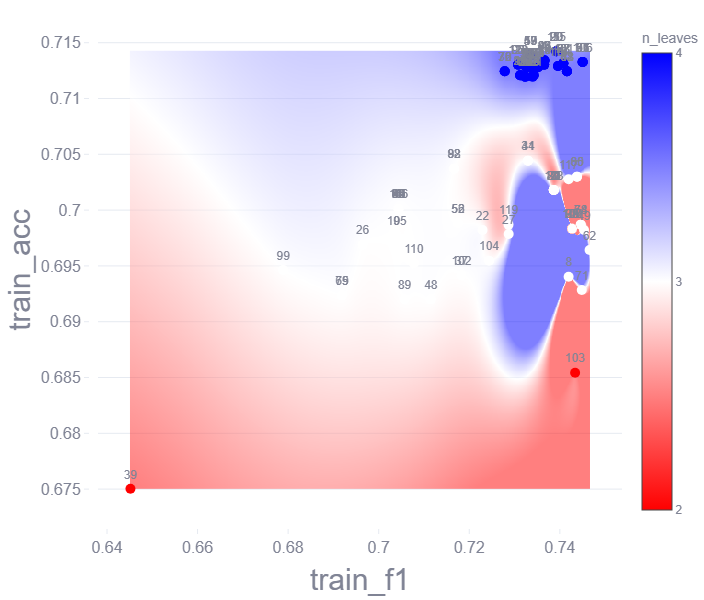}
  \includegraphics[scale=0.3]{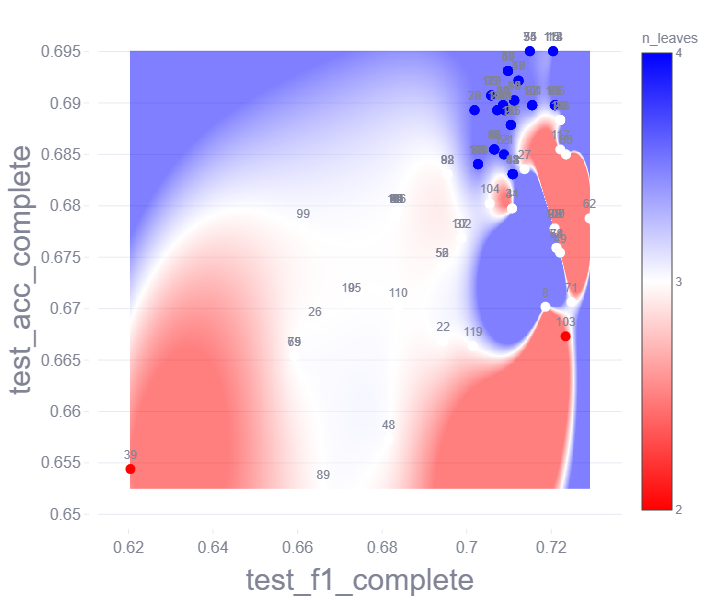}
  \caption{Comparison of the FICO dataset train and test performance. The color represents the number of leaf nodes in a decision tree model.}
  \label{fig:FICO_performance}
\end{figure}

\subsection{Rashomon Set based on FICO Dataset}

The FICO dataset refers to a public dataset provided by FICO (Fair Isaac Corporation) for credit scoring modeling. It is used to develop and evaluate ML models for predicting individual credit risk. 


We collected a Rashomon set of ML models from the FICO dataset using the TreeFarms method. The TreeFarms implementation can be found at https://github.com/ubc-systopia/treeFarms. TreeFarms is specifically designed to generate sets of nearly optimal decision trees, making it suitable for exploring the Rashomon set in interpretable ML contexts. The framework constructs multiple decision trees that achieve similar performance while maintaining different structural characteristics, allowing us to study the diversity of possible explanations for the same prediction task~\cite{yang:2017:scalable}.


The TreeFarms configuration includes several parameters that control the model generation process. In the preprocessing stage, we employed GOSDT's (Generalized Optimal Sparse Decision Tree)~\cite{hayden:2022:fast} threshold guessing method to binarize the data, with "n\_est=40" and "max\_depth=1" for threshold generation. 


To handle missing values, which are common in real-world financial data like FICO, we implemented the "nansafe\_cut" function. This function extends the traditional binary splitting approach by explicitly handling N/A values, treating them as a separate category in the decision process. Specifically, when a feature value is missing, the "nansafe\_cut" function applies a logical OR operation between the standard threshold condition and the N/A check, ensuring that missing values are handled consistently across all decision boundaries. All features are converted into binary features in the form of "feature\_name$<=$threshold", where the threshold values are determined through GOSDT's optimization process.


This binarization step is necessary for several reasons: First, it transforms the continuous feature space into a discrete one, making the decision boundaries more interpretable while maintaining the predictive power of the features. Second, the binary representation ensures consistency with GOSDT's theoretical guarantees about optimality, as the algorithm was designed to work with binary decision rules. Finally, this transformation simplifies the subsequent tree construction process in TreeFarms while preserving the meaningful patterns discovered during the GOSDT threshold optimization phase.


In the FICO dataset, we analyzed the model collection from the Rashomon set using the VAR. For each model, we calculated the accuracy and F1 scores on both training and testing sets, while also recording the number of leaf nodes and training loss. 


Through visualization analysis of the results, in Fig.~\ref{fig:FICO_performance} we observed that it is commonly known that shallower decision tree models represent simpler tree structures and should have lower prediction accuracy. This perception aligns with the expected results of the training set, as evident from the training accuracy and F1 scores. However, when examining the models on the testing set (test\_acc\_complete and test\_f1\_complete), we found that some "good" models drop a lot in the testing results. It needs to be noted that this differs from the general understanding that a model's performance on training data differs from its performance on test data. This is because while the Rashomon set has collected models that perform similarly on test data during the collection process, there is still a lack of support for model selection under certain feature constraints or conditions. This finding can help ML developers screen for appropriate models under specific metrics or feature conditions.


As shown in Fig.~\ref{fig:FICO_performance}, shallower trees(in red) with fewer leaf nodes can still yield good performance on test accuracy and F1 scores. This visual insight underscores the point about variability in performance across models of different complexities. The RBF-heatmap plot shows a heatmap representation of model performance on testing data, where colors indicate different levels of performance, reflecting the impact of the number of leaf nodes on test accuracy and F1 scores. With the VAR analysis system, we are able to effectively compare differences across multiple models visually and help the ML model developers compare and select the targeted models. 


\begin{figure}[!ht]
  \centering
  \includegraphics[scale=0.3]{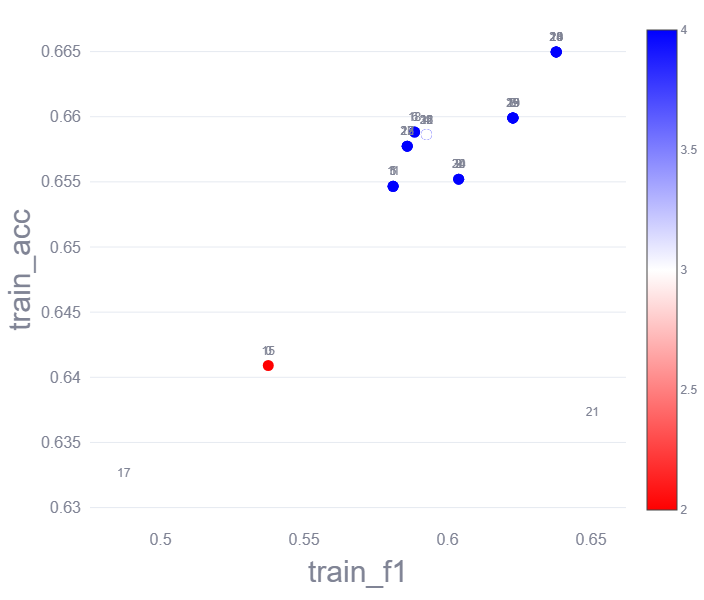}
  \includegraphics[scale=0.3]{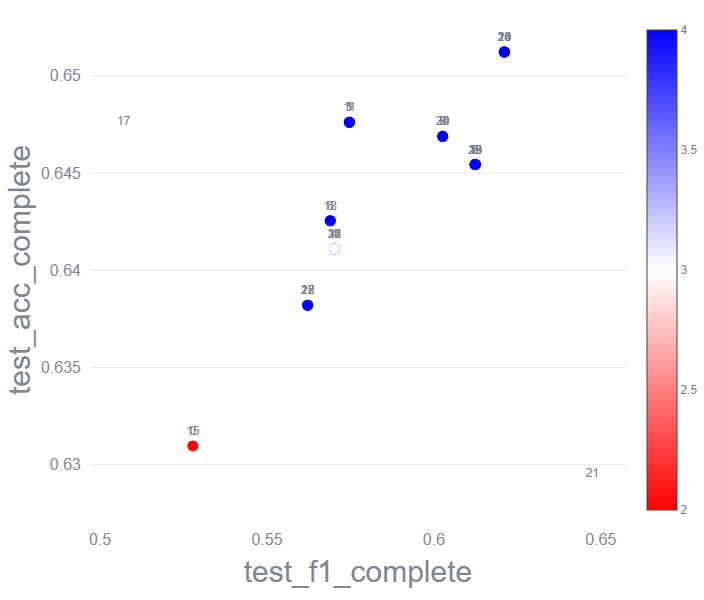}
  \caption{Comparison of the COMPAS dataset train and test performance. The color represents the number of leaf nodes in a decision tree model.}
  \label{binned_performance}
\end{figure}

\subsection{Rashomon Set based on COMPAS Dataset}

To assist experts and developers using TreeFarms to collect Rashomon sets, VAR also leveraged another public dataset, the COMPAS dataset. Unlike the visualization and interaction of tree structures in TimberTrek, VAR focuses on the comparison of the collected multiple models. Developers, after becoming familiar with TreeFarms for the Rashomon set collection, can transition to using the practical VAR approach.


The COMPAS (Correctional Offender Management Profiling for Alternative Sanctions) dataset is a widely used public dataset in the field of criminal justice and recidivism prediction. It was originally compiled and analyzed by ProPublica, a nonprofit investigative journalism organization, to examine racial bias in algorithmic risk assessment tools used in the criminal justice system. The COMPAS dataset contains information on 6,907 criminal defendants from Broward County, Florida. Each record represents an individual defendant and includes 7 continuous features such as the defendant's age, number of prior convictions, number of juvenile felonies, and COMPAS risk scores. The dataset was designed to predict the likelihood of a defendant recidivating, or committing another crime after being released. 


In the COMPAS dataset, we used the VAR system following steps similar to those used in collecting the Rashomon set from the FICO dataset. For each model, we calculated accuracy and F1 scores on both training and testing sets, while also recording the number of leaf nodes and training loss. Through visualization analysis of the results, we observed a phenomenon similar to before: models that performed well on training data did not always show consistent performance on test data. When selecting models with "similar" performance, the Rashomon set evaluates models based on multiple metrics. However, when developers need to find model parameters under a specific criterion, they can use VAR to quickly screen for suitable models.


As shown in Fig.~\ref{binned_performance}, we obtained similar findings by collecting the Rashomon set from the COMPAS dataset. In the COMPAS dataset, simpler decision tree models (with fewer leaf nodes) can achieve comparable performance to complex models on the testing set. This further validates our previous observation: model complexity does not always show a positive correlation with its performance in a Rashomon set. This finding suggests that simpler and more interpretable models can still maintain a reasonably good performance. It needs to be noted that this conclusion is not the focus of this work. We hope to use these two public datasets as examples to illustrate the role of VAR in analyzing and comparing several ML models in a Rashomon set. 


Unlike traditional scatter plots, the RBF-dot mode provides a third dimension for data analysis with the introduction of the color axis. This visual plot intuitively helps researchers observe potential patterns and correlations in datasets, offering a convenient approach to multidimensional data analysis. Through color variations, users can quickly identify the distribution characteristics of data points, explore relationships between different dimensions, and thus gain deeper insights compared to traditional scatter plots. Unlike heatmaps, where colors represent regional aggregation features, color assignment in scatter plots is specific to individual data points. This means that the color of each point carries a unique information dimension, precisely reflecting the specific characteristics of that data point in a multidimensional space.


Through two case studies on public datasets, we can see that the VAR has demonstrated good practicality in analyzing and comparing Rashomon sets. The system not only effectively displays differences in model performance across various evaluation metrics but also helps users better understand the trade-offs involved in model selection. This multidimensional comparative analysis approach provides more comprehensive decision support for model selection in practical applications.


\section{User Evaluation}
\label{sec:evaluation}
\subsection{User Background}

To further validate the effectiveness of VAR, we invited five senior experts from O University and Z Institution to evaluate the VAR in analyzing and comparing ML models in the Rashomon set. The experts compared the VAR with their previously used traditional numerical analysis methods to identify its advantages and shortcomings in the model selection process.


The three experts from O University have research backgrounds in computer science and visualization, focusing on data science, model interpretability, and visual analysis, with rich technical experience and analytical capabilities. The two experts from Z Institution come from the fields of educational data analysis, historical data reconstruction, and visualization, bringing an interdisciplinary research perspective and broad data analysis capabilities. These experts have accumulated years of professional experience in their respective fields and are familiar with data-driven analysis methods.


Before using the VAR, except for the experts with visualization backgrounds, most experts were accustomed to using numerical analysis and statistical methods to evaluate and compare model performance, often relying on tabular data and quantitative indicators, and rarely utilizing visualization tools for more intuitive model analysis. The traditional approaches have limitations in identifying subtle differences in models and providing interpretability, especially when dealing with large, multi-dimensional datasets, where it is often difficult to demonstrate the complex relationship structure effectively. By introducing the VAR, we provide these experts with a more intuitive and interactive analysis approach compared with the traditional numerical analysis methods.


The feedback from the expert users, with their high level of professionalism and representativeness, can provide profound insights and a basis for improving the VAR in actual applications, thereby helping us to optimize the system's practicality and interpretability in multi-disciplinary contexts.


\subsection{User Feedback}

During the evaluation process, the experts first used their customary numerical analysis methods to screen the models in the Rashomon set, which served as the benchmark. Subsequently, they used VAR's interactive heatmaps and colored-dot plots to screen the models, to compare the differences and pros and cons of the two analysis approaches. By collecting the experts' feedback after using the system, we gathered the following content and conducted relevant organization and summarization. The following are some of the raw feedback from the experts, with the italics representing the original comments.


Expert A: \textit{"The system provides an intuitive understanding of data distribution at both macro and micro levels. Especially in complex model data, the color gradients highlight important patterns, saving me considerable time."}

Expert B: \textit{"The system is highly interactive. Through slider bars and color selectors, I can quickly adjust parameters and observe model performance under different settings. This often requires additional data processing steps in other tools."} 

Expert C: \textit{"During model comparison, I can discover phenomena that are typically difficult to observe intuitively. For example, color gradients in densely distributed areas reveal potential patterns and outliers, helping me identify possible model deficiencies in specific scenarios."}

Expert D: \textit{"The system's noise handling functionality allows me to see more realistically how models perform under different conditions, especially when data has subtle variations, helping me identify more robust models."}

Expert E: \textit{"The diversity of RBF interpolation options is very practical, allowing me to switch between different interpolation methods and understand the similarities and differences between models. This kind of multi-dimensional analysis capability is rare in previous tools."}

Through expert feedback, we can find that VAR provides intuitiveness and flexibility to the experts. Experts indicated that the file upload functions were easy to operate, allowing quick switching between datasets, and making it suitable for comparative analysis of different models. Regarding slider bars and color scale selection, the interactive controls met their needs in adjusting resolution, selecting different RBF functions, and changing color ranges. The system received recognition for its smooth interaction in switching between model heatmaps and scatter plots and selecting feature axes. Through the overlay of RBF interpolation heatmaps with model scatter plots, experts could more quickly identify differences between models, and the data structure visualization enhanced decision-making efficiency.


Experts also mentioned that, compared with existing tools, such as \textit{TimberTrek} and \textit{GamChanger}, focus on specific aspects of model exploration. For example, TimberTrek provides a sunburst-style visualization to examine the structural properties of decision trees, aiding users in identifying shared features and decision pathways. Similarly, GamChanger facilitates interactive exploration, allowing users to edit and understand model parameters intuitively. While these tools excel in their respective domains, they are often constrained to a specific model type or lack the ability to perform horizontal comparisons across multiple models within a Rashomon set.


In contrast, VAR is designed to analyze the Rashomon set as a whole, providing both global and local insights into model performance under varying feature conditions. Its dual visualization RBF-heatmap mode and RBF-dot mode enable users to explore performance trends at a level while supporting precise, model-specific analysis. This combination is effective for understanding the diversity and robustness of models, as highlighted by experts during the evaluation process.


According to expert feedback, VAR addresses several limitations of existing tools. One expert noted, \textit{"The ability to switch between heatmap and scatter plot modes allows me to explore both global trends and individual model details, which is often missing in other tools."} Another expert emphasized that \textit{"the RBF-based interpolation provides a smooth visualization of performance patterns across the feature space, helping me identify regions where models might be underperforming or excelling."} These comments highlight the flexibility of VAR in accommodating both exploratory and targeted analyses, which is less feasible with tools that focus solely on structural properties or single-model interpretability.


Furthermore, VAR introduces enhanced interactivity, allowing users to filter models, adjust visualization parameters, and examine missing data conditions directly within the interface. Compared to traditional numerical analysis methods or tools like TimberTrek, which require manual calculations or fixed visualizations, VAR streamlines the analytical workflow, reducing the time and effort needed for model screening. One expert from the evaluation stated, \textit{"VAR simplifies the process of identifying robust models under specific feature constraints, which would otherwise require multiple iterations of manual computation."}


In addition to its usability, VAR is also adaptable across different ML contexts, as demonstrated in the case studies involving the FICO and COMPAS datasets. By integrating RBF-based visualizations, it effectively bridges the gap between global trends and granular model behavior, providing a unified framework for comparing models across diverse application scenarios.


The feedback from experts provides multiple levels of insight for future optimization of the VAR system. The current strength of the system lies in its combination of intuitiveness and flexibility, providing ML developers who analyze the Rashomon set with new perspectives on the collected models.


\subsection{Limitation and Discussion}

Based on the feedback from experts and users, it is acknowledged that the design of VAR is not perfect. The current VAR has some limitations according to the expert feedback. Some experts suggested providing more data preprocessing guidance and instructions, hoping that the interactive visualizations in VAR can better help users understand the impact of parameter settings on the data, such as providing preview images or thumbnails. Some experts also expressed the need for a multi-heatmap comparison function because VAR only supports single-image download, and users hope to be able to more conveniently compare multiple heatmap results with different parameter settings on the interface, to observe the differences between models more accurately. 


The most mentioned concern is the issue of point overlap. Since the original RBF-heatmap mode processes a large amount of model data, ML model developers provided feedback that a subset of data points is needed to complete their tasks. This insight helped avoid issues in the RBF-dot mode, where the size of the dots affects the number of points that can be displayed in the graph. Based on the principle of filtering, if points overlap, the impact on the values of the points themselves is considered from the perspective of RBF mathematical calculations, typically reflected in a darker color and a higher value. Additionally, in the actual user interface, there is a tick box that can toggle the index on and off. When enabled, users can observe overlapping indices, which helps ML model developers identify overlapping areas to some extent.


To effectively demonstrate the practical applications and analytical capabilities of RBF-based visualization patterns, we selected FICO and COMPAS datasets for two case studies in Section~\ref{sec:case_study}. The FICO dataset focuses on predicting credit default risk, encompassing a wide range of feature combinations and numerous models in the Rashomon set. Given the need to identify overall performance trends in the feature space (such as areas with high-performing model clusters or significant performance variations), the RBF-heatmap pattern is more suitable compared to the RBF-dot pattern. It provides a global perspective that intuitively explores model diversity and robustness, aligning with the dataset's high-dimensional nature and the goal of discovering broader patterns rather than focusing on individual models.


The COMPAS dataset used for recidivism prediction presents different analytical challenges. Its relatively smaller feature space and the need for precise model-specific analysis make the RBF-dot pattern more appropriate. By visualizing individual models as discrete points, this pattern avoids interpolation and enables detailed examination of specific models. Tasks such as identifying best-performing models under specific feature constraints or exploring outliers in model performance benefit greatly from this level of granularity. This choice is valuable for exploring fairness-related issues, which is a key aspect of the COMPAS dataset.


While the selected modes have proven effective for their respective case studies, we acknowledge that both modes are complementary and can be tailored to different research scenarios.


Given that the current VAR is still in its early development stage, we plan to further enrich its functionalities in the future to better meet the needs of ML model developers. To this end, we will continue to expand the user base of VAR, encouraging more ML model developers to use VAR to compare and analyze the differences between similar ML models. After the authors from O University introduced VAR to the ML developers at D University, the developers at D University plan to conduct more extensive research with a user group of about 20 people. By collecting feedback from different domains, we will optimize the system's user experience.

\section{Conclusion}

In this paper, we propose VAR, a visual analytics solution for comparing differences in ML models with similar performance within the Rashomon set across various ML application scenarios. Through application on public datasets and user evaluation, we demonstrate that VAR, utilizing visualization techniques, can effectively assist ML model developers in screening suitable models for tasks with different requirements.


To the best of our knowledge, this approach has not been previously reported in the literature. We will continue to develop this work, primarily focusing on further optimization of the web-based version of the software based on this method. We will further improve the functionality of the web version through collected user feedback. 


We anticipate that there are certainly many other visual patterns that can be used to compare ML models in the Rashomon set, and there are undoubtedly other ways to help quantify and compare the specific differences between these models. VAR offers a new perspective on the model selection process through visualization, helping ML model developers evaluate models from a more intuitive standpoint. We believe that the more ML model developers use VAR to analyze and compare differences between multiple similar ML models, the more potentially valuable methods will be proposed.






 


\bibliography{mybibfile}


\newpage
\begin{center}
\large
APPENDICES OF\\[1mm]
\Large\noindent
\textbf{\textsf{VAR: Visual Analysis for Rashomon Set of Machine Learning Models' Performance}}\\[2mm]
\normalsize
Yuanzhe Jin

\normalsize
\end{center}


\section{Parameter Settings in the Experiment}

In the FICO dataset, the depth budget (depth\_budget) is set to 4, which constrains the maximum depth of generated trees and ensures interpretability while maintaining reasonable model complexity. The Rashomon bound adder (rashomon\_bound\_adder) is set to 0.03, determining the acceptable performance gap from the optimal solution - this means we include trees whose performance is within 3\% of the best-performing tree. The regularization parameter (regularization) is set to 0.02, which helps balance model complexity with performance by penalizing the number of leaves in the trees. Additional configuration parameters include "rashomon\_bound\_multiplier" set to 0 and "trivial\_extensions" set to True, which help control the diversity of the generated tree set while avoiding redundant model structures. Through these parameter settings, we collected a Rashomon set consisting of 152 models from the FICO dataset.


As the COMPAS dataset is relatively simpler compared to the FICO data, the “depth\_budget” is set to 5, allowing for greater tree depth. The remaining parameters include the ”rashomon\_bound\_adder” set to 0.03, regularization set to 0.02, ”rashomon\_bound\_multiplier” set to 0, and “trivial\_extensions” set to True. Through these parameter settings, we collected a Rashomon set consisting of 32 models from the COMPAS dataset.


\section{Comparison of the different kernel functions for the RBF heatmap visualization}

We present 16 different types of kernel functions using the same data points. Four are in each cluster with the mathematical format above the Figure. In all the kernel functions below, $r$ represents radial distance, $\sigma$ and $c = 1.0$ are kernel function parameters.

By comparing the visualizations of these 16 kernel functions, we can gain a better understanding of their characteristics in RBF heatmaps, providing valuable insights for selecting the appropriate kernel function for specific application scenarios. Overall, the classical kernel functions in the first group perform the most consistently, while custom kernel functions may deliver superior results in specialized contexts. 

\subsection{Group 1: Classical Kernel Functions}

\begin{figure}[!ht]
  \centering
  \begin{tabular}{cc}
    \includegraphics[scale=0.30]{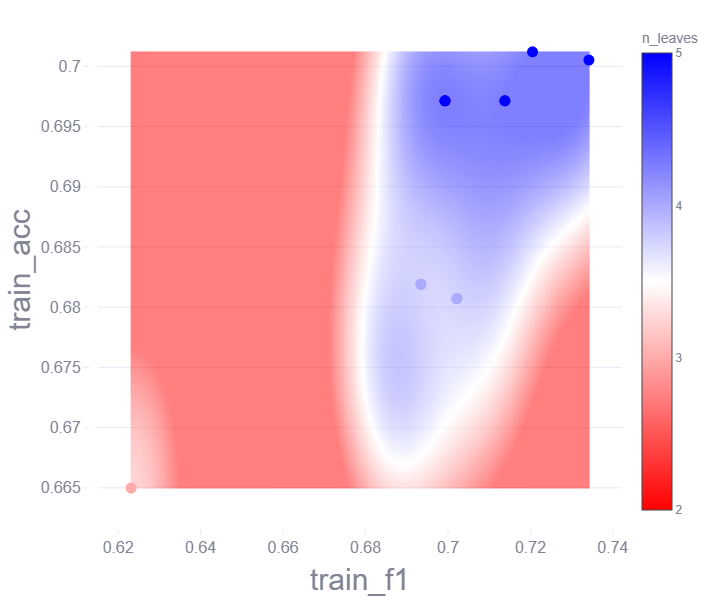} &
    \includegraphics[scale=0.30]{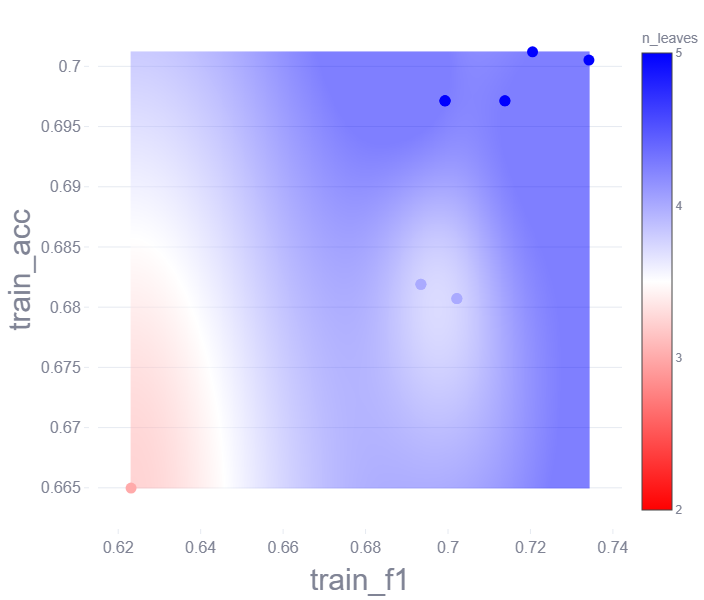} \\
    \includegraphics[scale=0.30]{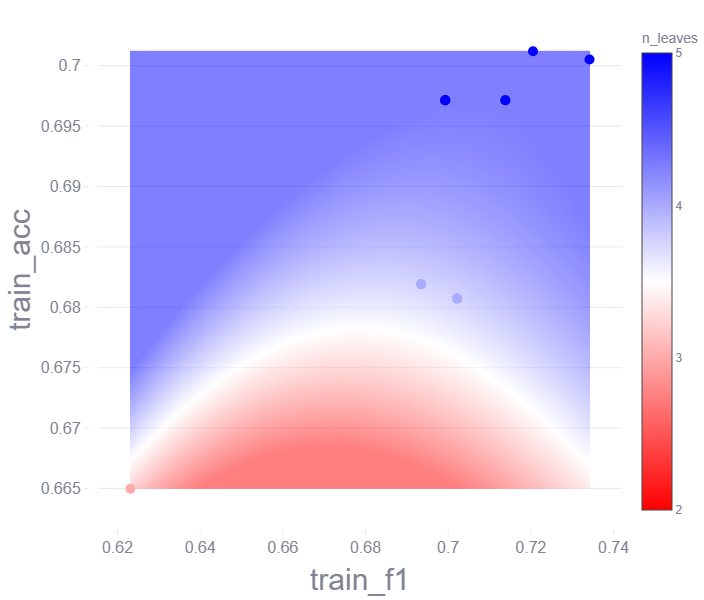} &
    \includegraphics[scale=0.30]{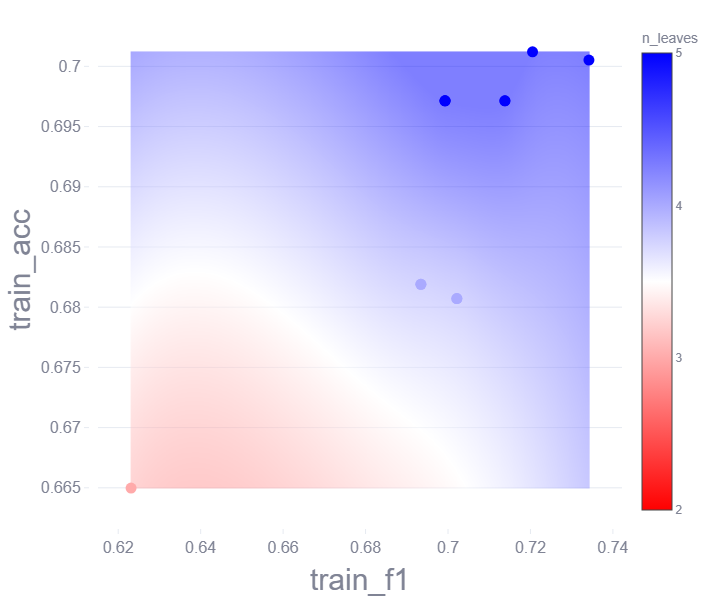}
  \end{tabular}
  \caption{Comparison of different kernel functions used in the RBF-heatmap mode: Gaussian (top left), Multiquadric (top right), Inverse Multiquadric (bottom left), and Thin Plate (bottom right).}
  \label{kernel_function_1}
\end{figure}

The first group shown in Fig.~\ref{kernel_function_1} includes the most commonly used classical kernel functions: Gaussian kernel, multiquadric kernel, inverse multiquadric kernel, and thin plate spline kernel. These kernel functions are widely applied in the field of machine learning and possess excellent mathematical properties. From the visualization results, it is evident that this group of kernel functions generally exhibits good smoothness and continuity, making them particularly suitable for capturing the overall distribution trends in data. The mathematical formats of the kernel functions in Group 1 are shown below:

\noindent Gaussian Kernel: $k(r) = \exp\left(-\frac{r^2}{2\sigma^2}\right)$

\noindent Multiquadric Kernel: $k(r) = \sqrt{r^2 + c^2}$

\noindent Inverse Multiquadric Kernel: $k(r) = \frac{1}{\sqrt{r^2 + c^2}}$

\noindent Thin Plate Kernel: $k(r) = r^2 \ln(r)$

\subsection{Group 2: Basic Polynomial Kernel Functions}

\begin{figure}[h] 
  \centering
  \begin{tabular}{cc}
    \includegraphics[scale=0.30]{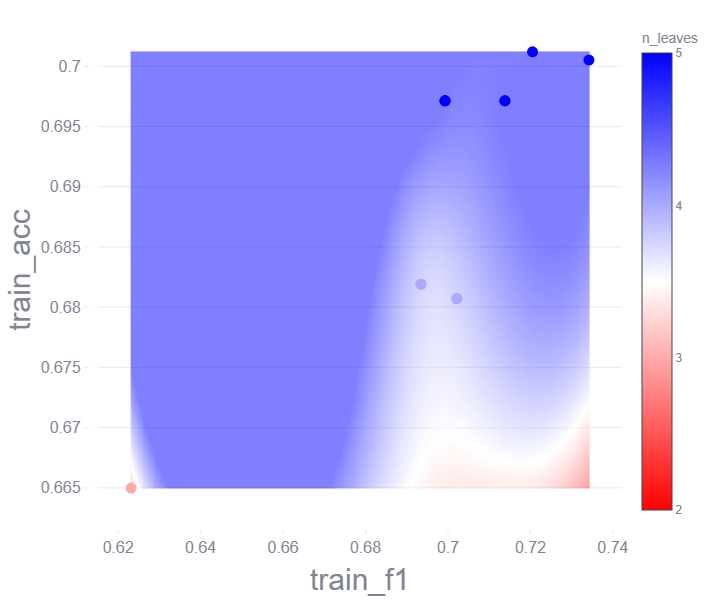} &
    \includegraphics[scale=0.30]{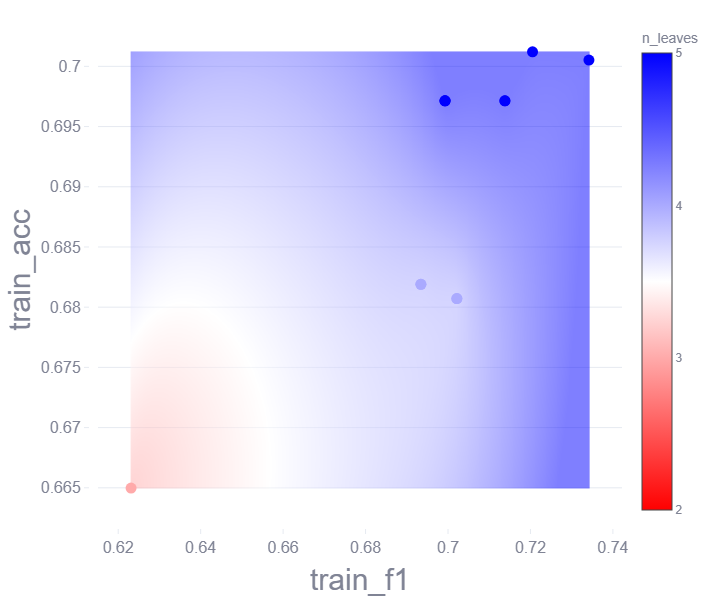} \\
    \includegraphics[scale=0.30]{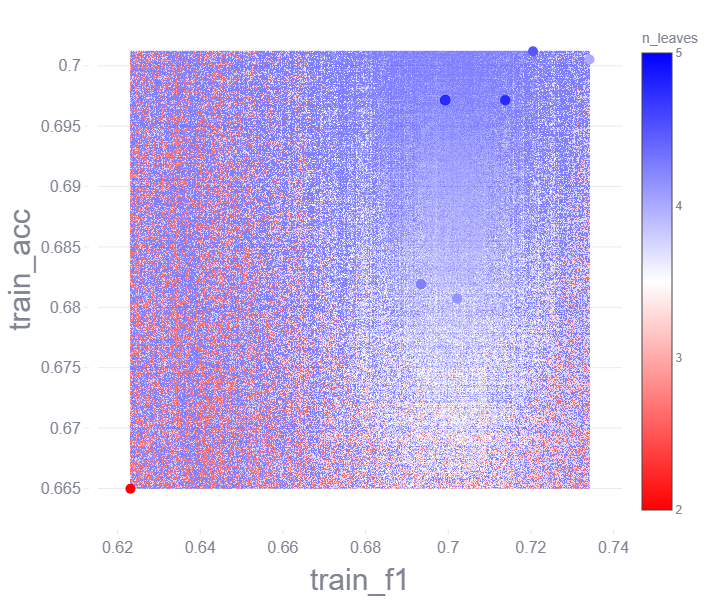} &
    \includegraphics[scale=0.30]{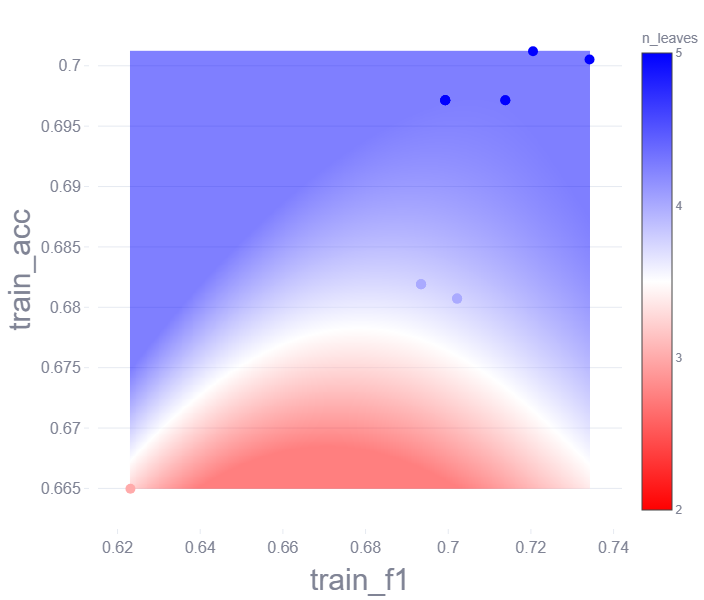}
  \end{tabular}
  \caption{Comparison of different kernel functions used in the RBF-heatmap mode: Cubic (top left), Linear (top right), Quadric (bottom left), and Inverse Quadric (bottom right).}
  \label{kernel_function_2}
\end{figure}

The second group shown in Fig.~\ref {kernel_function_2} is mainly composed of basic polynomial functions, including cubic kernels, linear kernels, quadratic kernels, and inverse quadratic kernels. These kernel functions are simple in form and computationally efficient, but they may not be as flexible as the first group when dealing with complex patterns. Observations from the heatmap show that this group of kernel functions provides a relatively direct interpolation effect, and the boundary features are more obvious. However, some kernel functions may not be able to handle some boundaries, resulting in noise points. The mathematical format of the kernel functions in the second group is as follows:

\noindent Cubic Kernel: $k(r) = r^3$

\noindent Linear Kernel: $k(r) = r$

\noindent Quadratic Kernel: $k(r) = r^2$

\noindent Inverse Quadratic Kernel: $k(r) = \frac{1}{r^2 + c^2}$

\subsection{Group 3: Mixed-Type Kernel Functions}

\begin{figure}[h] 
  \centering
  \begin{tabular}{cc}
    \includegraphics[scale=0.30]{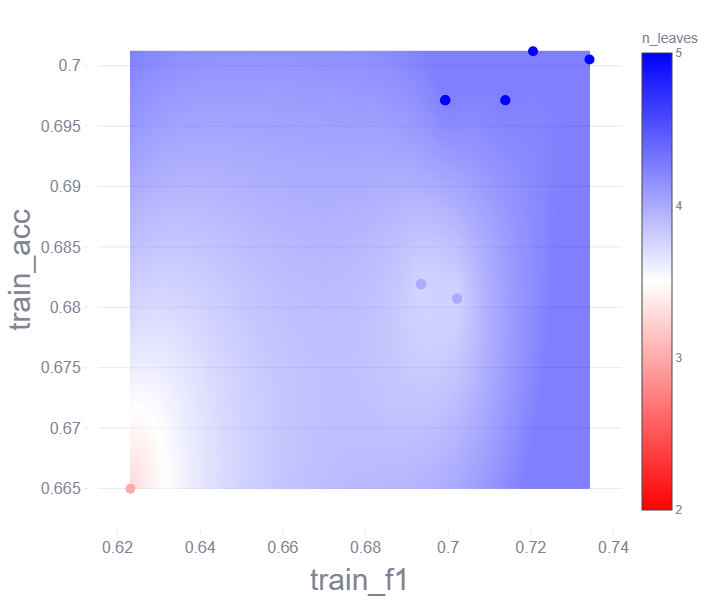} &
    \includegraphics[scale=0.30]{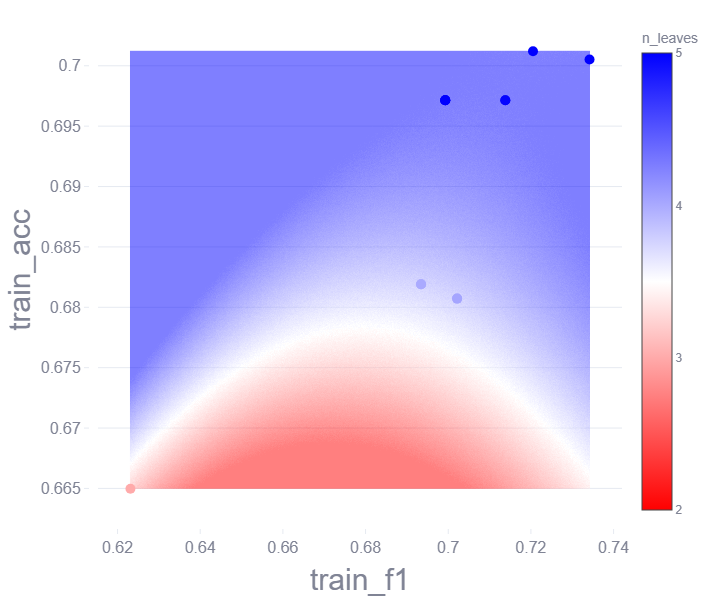} \\
    \includegraphics[scale=0.30]{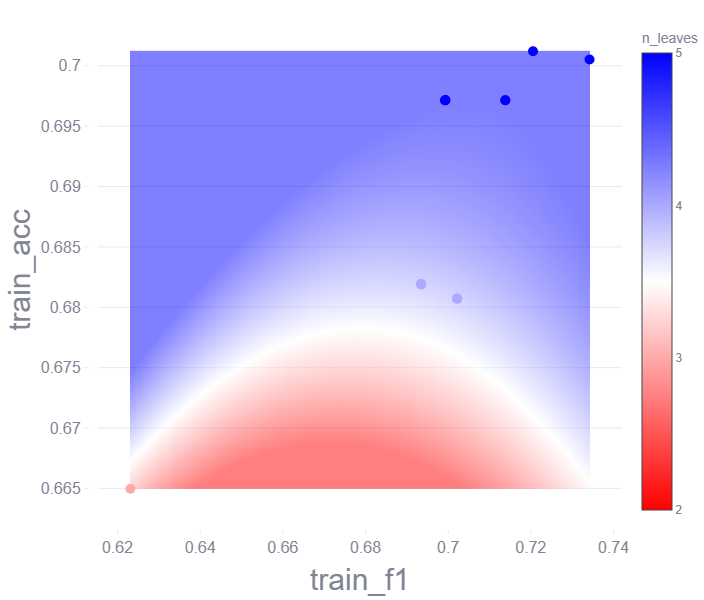} &
    \includegraphics[scale=0.30]{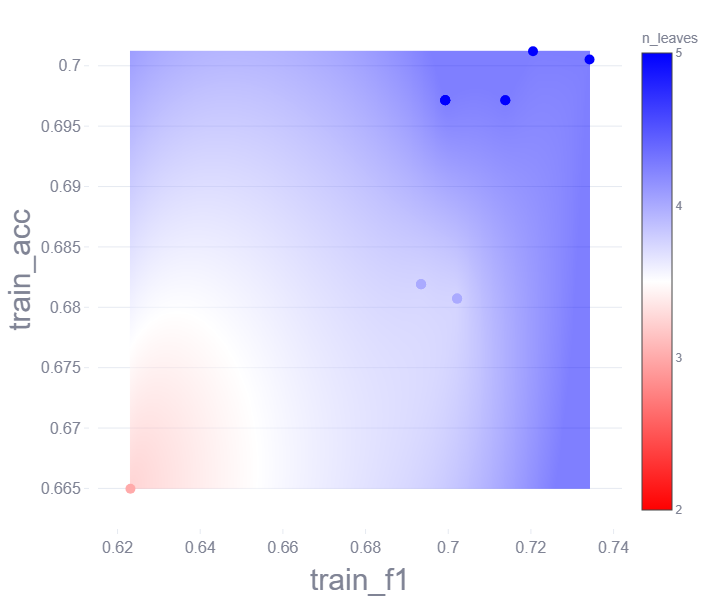}
  \end{tabular}
  \caption{Comparison of different kernel functions used in the RBF-heatmap mode: Spine (top left), Beckmann (top right), Wave (bottom left), and Logarithmic (bottom right).}
  \label{kernel_function_3}
\end{figure}

The third group shown in Fig.~\ref{kernel_function_3} contains the spline kernel, the Beckmann kernel, the wave kernel, and the logarithmic kernel, which combine different mathematical properties. Visualizations show that this group excels in expressing local details, each with its unique ability to capture different patterns in the data. The mathematical formats of the kernel functions in Group 3 are shown below:

\noindent Spline Kernel: $k(r) = r \ln(r)$

\noindent Beckmann Kernel: $k(r) = \exp\left(-\frac{r^2}{2c^2}\right)$

\noindent Wave Kernel: $k(r) = \frac{\sin(r)}{r}$

\noindent Logarithmic Kernel: $k(r) = \ln(r + 1)$

\subsection{Group 4: Custom Kernel Functions}

\begin{figure}[h] 
  \centering
  \begin{tabular}{cc}
    \includegraphics[scale=0.30]{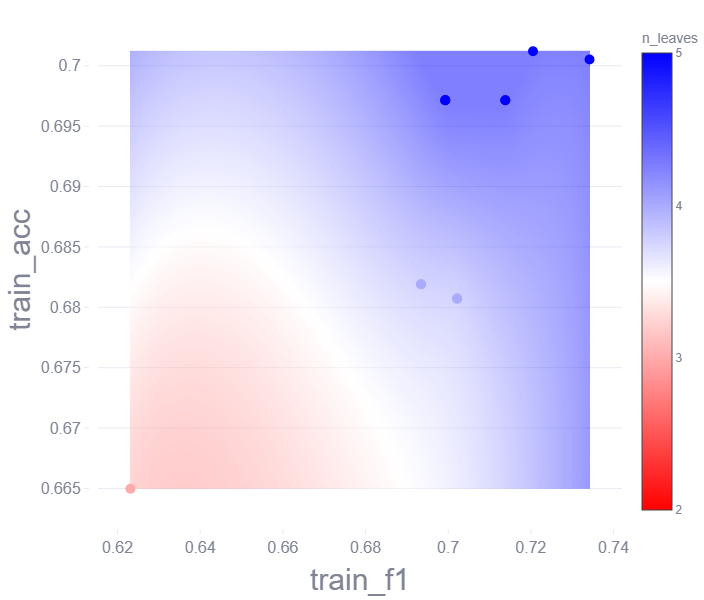} &
    \includegraphics[scale=0.30]{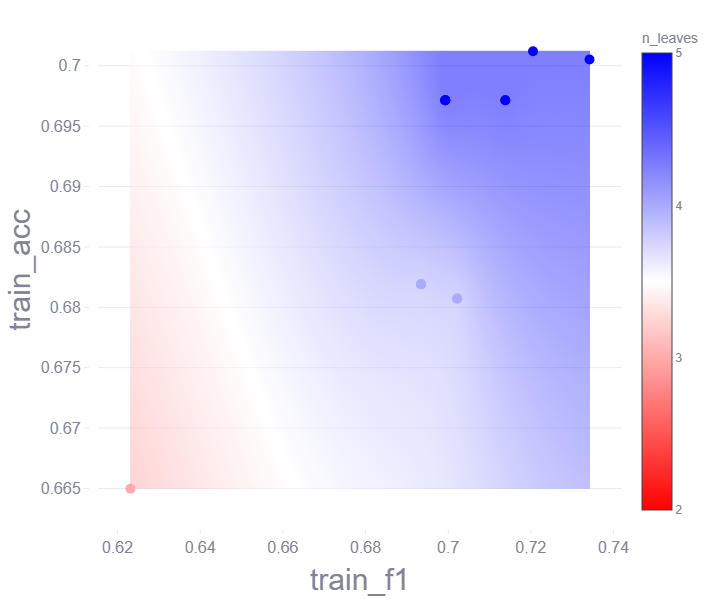} \\
    \includegraphics[scale=0.30]{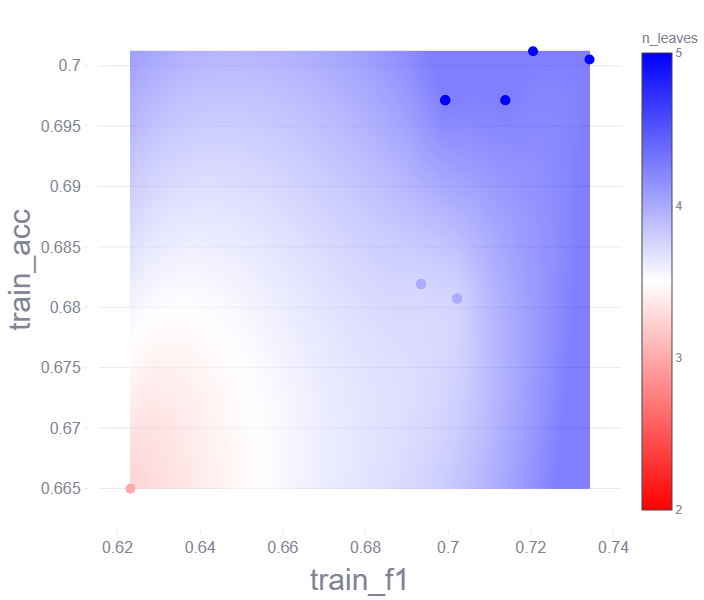} &
    \includegraphics[scale=0.30]{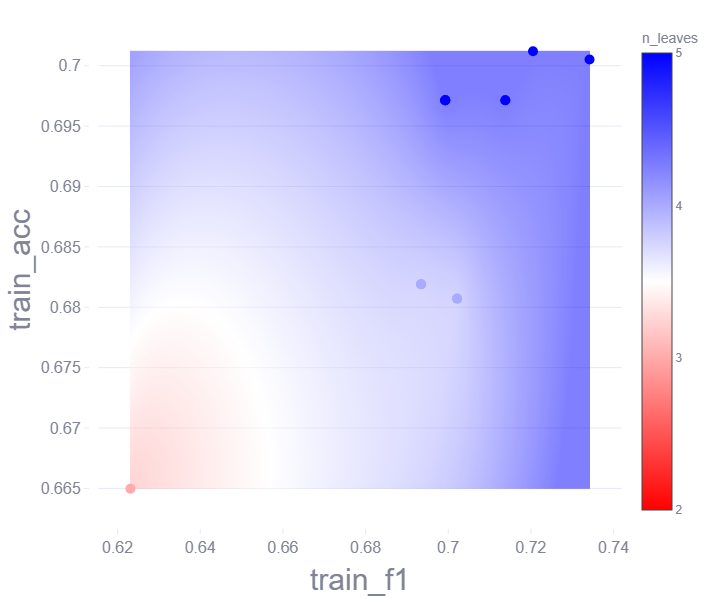}
  \end{tabular}
  \caption{Comparison of different kernel functions used in the RBF-heatmap mode: Paper used (top left), Exponential Root Kernel (top right), Sine Logarithmic Kernel (bottom left), and Hyperbolic Polynomial Kernel (bottom right).}
  \label{kernel_function_4}
\end{figure}

The final group shown in Fig.~\ref{kernel_function_4} comprises specially designed kernel functions, including those used in specific research papers and several composite kernel functions. By combining multiple mathematical properties, these kernel functions aim to achieve better performance in specific application scenarios. Heatmaps reveal that these functions can maintain overall smoothness while highlighting local features. The mathematical formats of the kernel functions in Group 4 are shown below:

\noindent The paper used kernel: $k(r) = \frac{r \cdot \log(1 + r^{0.5})}{1 + r^{0.1}}$

\noindent  Exponential Root Kernel:
$k(r) = \frac{\exp(-r) \cdot \sqrt{r + 1}}{1 + r}$

\noindent Sine Logarithmic Kernel:
$k(r) = \frac{\sin(r) + \log(1 + r)}{1 + r^2}$

\noindent Hyperbolic Polynomial Kernel:
$k(r) = \frac{arctanh(\tanh(r)) + r^{1.5}}{1 + r^{0.5} + r^3}$

\newpage
\section{High Resolution Images}

\begin{figure*}[ht]
  \centering
  \includegraphics[scale=0.48]{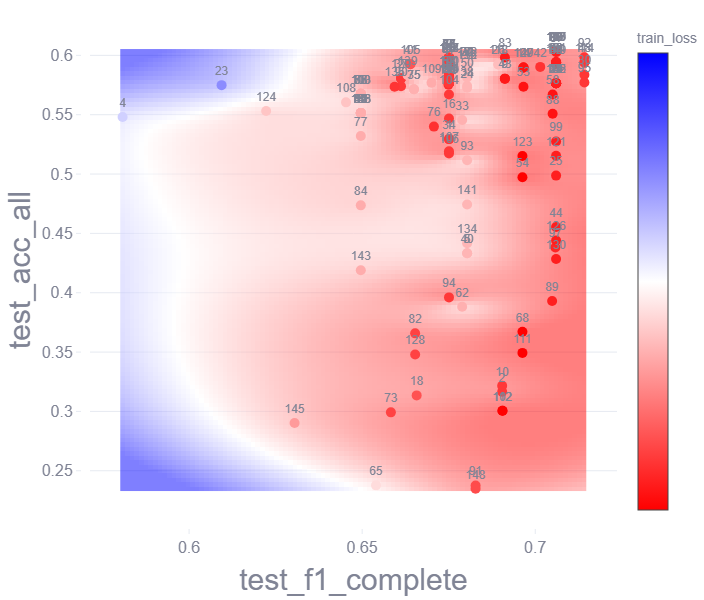}
  \includegraphics[scale=0.48]{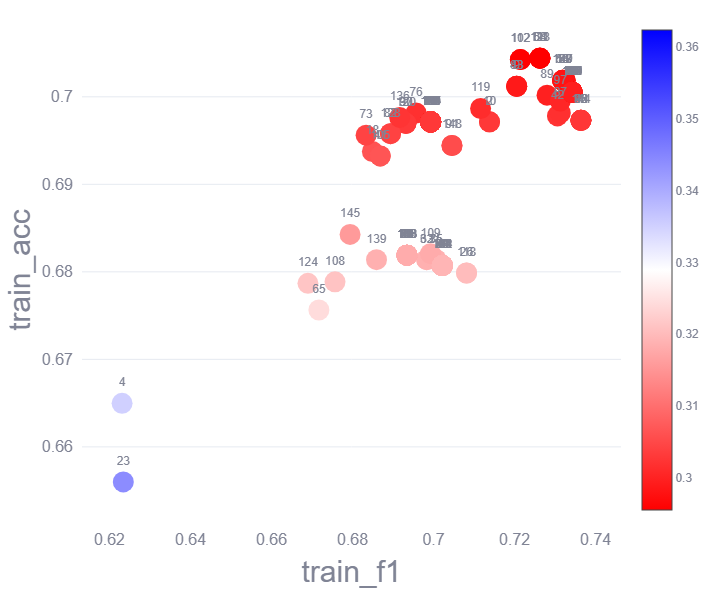}
  \caption{(a) RBF-heatmap mode (Left). Comparison of the performance of 152 models in the Rashomon set on the test set. To ensure that the color of the dots does not completely blend with the background, the color of the dots has been darkened. The color represents the train loss. (b) RBF-dot mode (Right). Comparison of the performance of the same 152 models on the training set. The color represents the train loss.}
\label{rbf_mode_high}
\end{figure*}

\begin{figure*}[ht]
  \centering
  \includegraphics[scale=0.56]{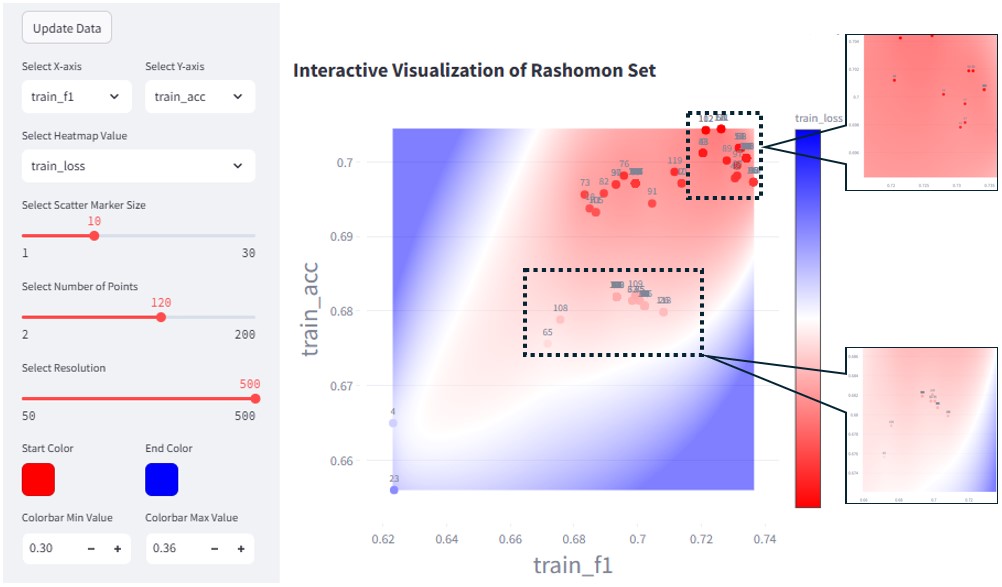}
  \caption{A screenshot of VAR showing functions in the control panel on the left and an RBF visualization plot on the right.}
\label{VAR_UI_high}
\end{figure*}

\begin{figure*}[ht]
  \centering
  \includegraphics[scale=0.48]{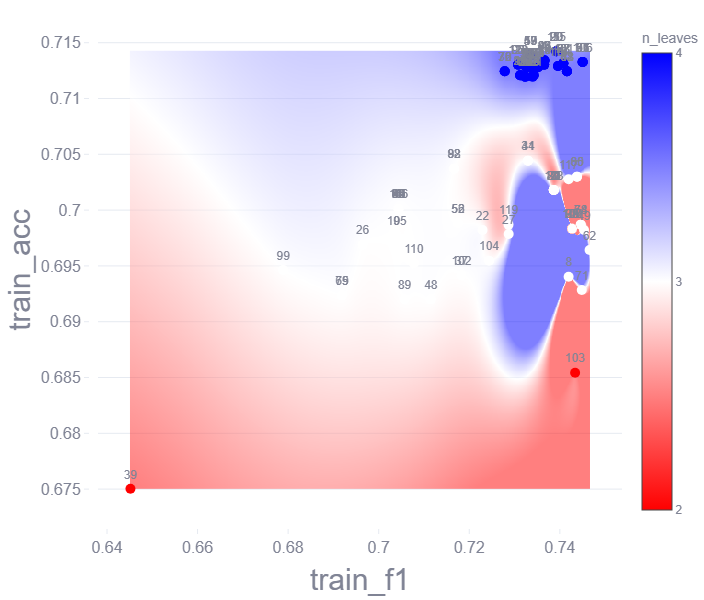}
  
  \includegraphics[scale=0.48]{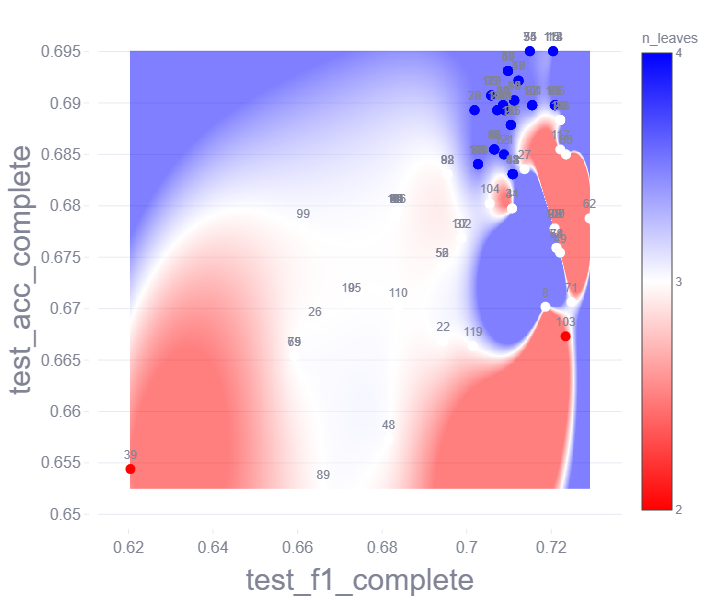}
  \caption{Comparison of the FICO dataset train and test performance. The color represents the number of leaf nodes in a decision tree model. Some models, like the label-89 model and the label-48 model, are special models that the ML model developers are trying to find.}
  \label{fig:FICO_performance_high}
\end{figure*}

\begin{figure*}[ht]
  \centering
  \includegraphics[scale=0.48]{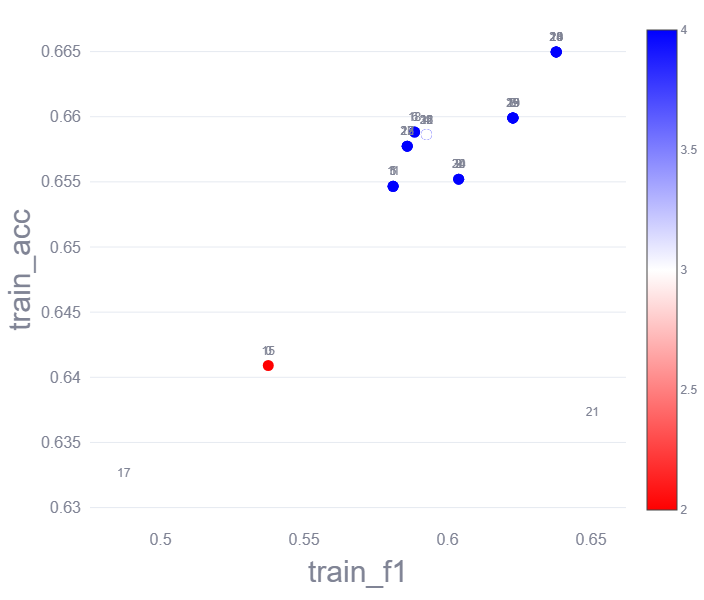}
  
  \includegraphics[scale=0.48]{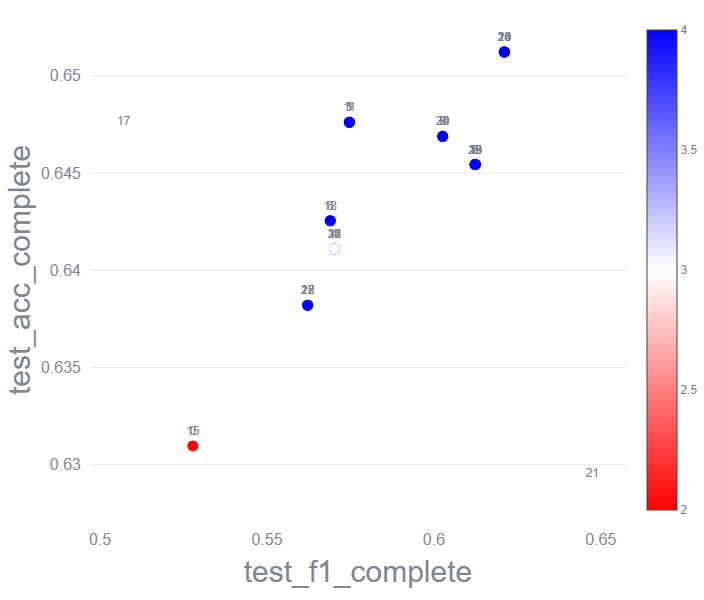}
  \caption{Comparison of the COMPAS dataset train and test performance. The color represents the number of leaf nodes in a decision tree model. The model labeled in 17 is the special model that the ML model developers are trying to find.}
  \label{binned_performance_high}
\end{figure*}

\end{document}